\documentclass[pdflatex,sn-mathphys-num]{sn-jnl}

\usepackage{graphicx}%
\usepackage{subcaption}
\usepackage{multirow}%
\usepackage{amsmath,amssymb,amsfonts}%
\usepackage{amsthm}%
\usepackage{mathrsfs}%
\usepackage[title]{appendix}%
\usepackage{xcolor}%
\usepackage{textcomp}%
\usepackage{manyfoot}%
\usepackage{booktabs}%
\usepackage{algorithm}%
\usepackage{algorithmicx}%
\usepackage{algpseudocode}%
\usepackage{listings}%
\usepackage{multirow} 
\usepackage{hyperref}
\usepackage{cleveref}

\theoremstyle{thmstyleone}%
\theoremstyle{thmstyletwo}%

\theoremstyle{thmstylethree}%

\begin{document}

\title[Article Title]{Autonomous Skeletal Landmark Localization towards Agentic C-Arm Control}

\author[1]{\fnm{Jay Hwasung} \sur{Jung}}\email{Jay-Hwasung.Jung@uvm.edu}

\author[1]{\fnm{Ahmad} \sur{Arrabi}}\email{Ahmad.Arrabi@uvm.edu}

\author[2]{\fnm{Jax} \sur{Luo}}\email{LUOJ2@ccf.org}

\author[2]{\fnm{Scott} \sur{Raymond}}\email{RAYMONS3@ccf.org}

\author[1]{\fnm{Safwan} \sur{Wshah}}\email{Safwan.Wshah@uvm.edu}

\affil[1]{\orgname{University of Vermont}, \orgaddress{\city{Burlington}, \postcode{05405}, \state{Vermont}, \country{U.S.A}}}

\affil[2]{\orgname{Cleveland Clinic}, \orgaddress{\city{Cleveland}, \postcode{44195}, \state{Ohio}, \country{U.S.A}}}

\abstract{

\textbf{Purpose}: Automated C-arm positioning ensures timely treatment in patients requiring emergent interventions. When a conventional Deep Learning (DL) approach for C-arm control fails, clinicians must revert to manual operation, resulting in additional delays. Consequently, an agentic C-arm control framework based on multimodal large language models (MLLMs) is highly desirable, as it can incorporate clinician feedback and use reasoning to make adjustments toward more accurate positioning. Skeletal landmark localization is essential for C-arm control, and we investigate adapting MLLMs for autonomous landmark localization. 

\textbf{Methods}: 
We used an annotated synthetic X-ray dataset and a real X-ray dataset. Each X-ray in both datasets is paired with several skeletal landmarks. We fine-tuned two MLLMs and tasked them with retrieving the closest landmarks from each X-ray. Quantitative evaluations of landmark localization were performed and compared against a leading DL approach. We further conducted qualitative experiments demonstrating: (1) how an MLLM can correct an initially incorrect prediction through reasoning, and (2) how the MLLM can sequentially navigate the C-arm toward a target location.

\textbf{Results}: On both datasets, fine-tuned MLLMs demonstrate competitive performance across all localization tasks when compared with the DL approach. In the qualitative experiments, the MLLMs provide evidence of reasoning and spatial awareness.

\textbf{Conclusion}: 
This study shows that fine-tuned MLLMs achieve accurate skeletal landmark localization and hold promise for agentic autonomous C-arm control. Our code is available at \url{https://github.com/marszzibros/C-arm-localization-LLMs.git}}


\maketitle

\section{Introduction}
\label{sec:introduction}

C-arm machines are widely used in complex interventions to provide real-time imaging guidance. However, their operation typically requires manual positioning to the region of interest (ROI) and coordination between multiple clinicians. In emergency procedures such as stroke thrombectomy, C-arms are often required overnight or on the weekend~\cite{late_time} and may be operated by less experienced staff in resource-limited settings~\cite{low-resource}. Prolonged C-arm positioning can delay treatment, compromise patient survival, and increase radiation exposure for both patients and clinical staff. Several efforts have explored deep learning (DL)–based automated C-arm control~\cite{c-arm-automation_first,c-arm-automation_second,ICCV_carm}. A major limitation of these approaches is their reliance on a fixed image-to-motion mapping; when a DL model fails, clinicians must revert to manual control, wasting valuable minutes that are critical for successful surgical outcomes. With the advent of agentic AI~\cite{MMedAgent} and multimodal large language models (MLLMs), a smart C-arm navigation system can function as an intelligent assistant capable of interpreting surgeon commands such as “align to the femoral neck” or “move to the lateral skull base.” When an initial prediction appears incorrect, the system can incorporate clinician feedback, reason over it, and autonomously adjust its output toward more accurate positioning.

Skeletal landmark localization on X-ray images provides stable anatomical reference points that define the patient’s orientation and geometry, which is central to C-arm positioning. By comparing the detected landmark configuration with target anatomical views, the navigation system can estimate the required C-arm motion to reach the target position. Existing work of DL–based approach for skeletal landmark localization~\cite{ISBI_carm} have performed well quantitatively; however, they operate purely on pixel-level supervision and lack the semantic capability to achieve context-aware, instruction-following, and interpretable localization that is needed for agentic C-arm control. Recent MLLMs, such as~\cite{MedGemma}, have demonstrated improved medical understanding compared to general-purpose pretrained LLMs. However, their expertise remains largely confined to chest X-rays and radiology report interpretation ~\cite{med_image_datasets,med_report_datasets}. 

\begin{figure}[!t]
  \centering
  \caption{Overview of the skeletal landmark localization and C-arm navigation framework. We first organize the datasets in the spatial grounding format, where each X-ray image is paired with several anatomical landmarks. The MLLM is then prompted to localize the three landmarks closest to the image center and trained using supervised fine-tuning with LoRA \cite{qlora}. Finally, the fine-tuned MLLM is tasked with localizing the closest landmark and navigating the C-arm toward a target location. The model outputs both the predictions and the corresponding reasoning. 
  }
  \label{fig:Fig1}
  \includegraphics[width=1.0\textwidth]{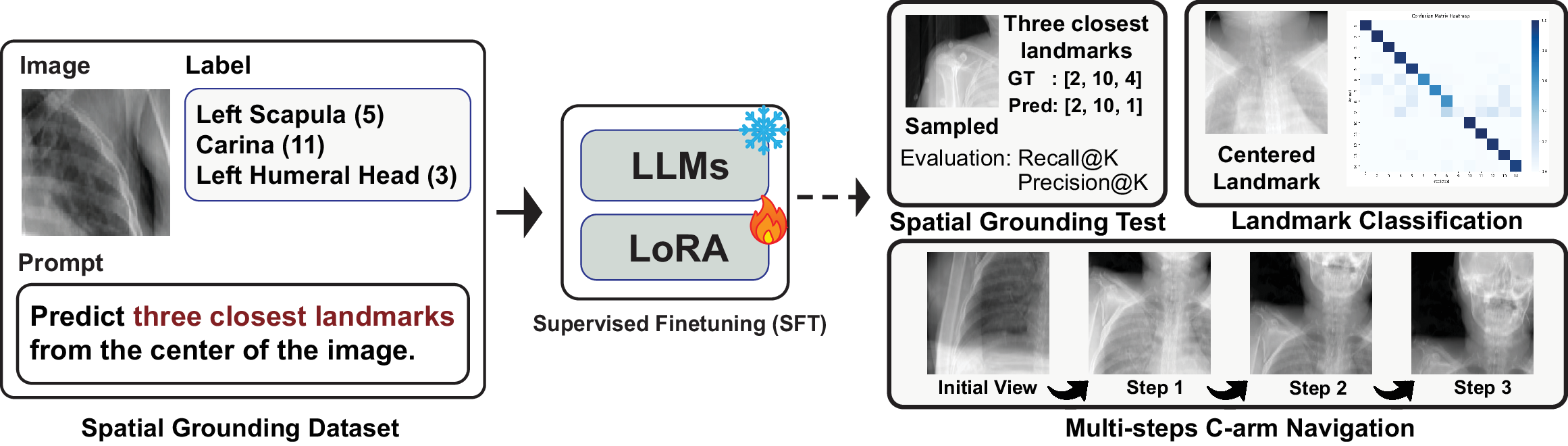}
\end{figure}

In this study, we investigate whether MLLMs can be further adapted to perform autonomous skeletal landmark localization. Grounded cognition theories~\cite{grounded_cognition} suggest that MLLMs can acquire spatial understanding by linking visual inputs with semantic concepts. Building on this insight, we propose a fine-tuning framework for X-ray skeletal landmark localization. This framework establishes anatomical spatial grounding by training MLLMs to predict nearby anatomical landmarks, enabling the model to infer spatial relationships among different anatomical structures. We conduct experiments on two datasets. The first is a public X-ray dataset \cite{unifesp}. The second is a synthetic dataset constructed by annotating a public upper-body CT dataset \cite{NMDID} with fourteen anatomical landmarks and generating corresponding Digitally Reconstructed Radiographs (DRRs) taking into account realism of C-arm imaging geometry. We argue that the synthetic dataset more faithfully reflects the clinical setting for two reasons: (1) prior work has demonstrated that AI models trained on synthetic DRRs generalize to unseen clinically acquired data without the need for re-training or domain adaptation~\cite{deepdrr}; and (2) the synthetic DRRs provide a field of view that more closely matches that of C-arm control compared with the real X-ray dataset \cite{unifesp}. We evaluate two open-source MLLMs, Gemma-3~\cite{gemma_2025} and Qwen-2.5VL~\cite{qwen2.5-VL}, and show that fine-tuned MLLMs  achieve competitive raw performance compared with a leading DL approach \cite{ISBI_carm}, and even out-perform it on several metrics. In qualitative experiments, we further provide clear evidence of the MLLM's reasoning and spatial awareness by demonstrating: (1) how an MLLM can correct an initially incorrect landmark prediction through reasoning, and (2) how the MLLM can navigate the C-arm toward a target location in a multi-step manner. Figure~\ref{fig:Fig1} gives an overview of the proposed framework. 

The remainder of this article is organized as follows. Section~\ref{sec:methods} provides a detailed formulation of landmark localization and C-arm control within the context of fine-tuned MLLMs. Section~\ref{sec:result} describes the synthetic data generation process, implementation details, and experimental evaluation of skeletal landmark localization and C-arm control. Finally, Section~\ref{sec:conclusion} discusses the limitations of this study and outlines directions for future work.

\section{Methodology}
\label{sec:methods}
The proposed framework is based on anatomical spatial grounding, which refers to the ability to localize an X-ray within the patient’s body relative to its anatomy. Supervised Fine-tuning (SFT) equips MLLMs with this positional understanding by learning relative relationships to nearby landmarks (i.e., the three closest), such as the skull being closer to T1 and the humeral heads than to the hemidiaphragm. This understanding intuitively supports downstream applications such as landmark classification and C-arm navigation.

\subsection{Skeletal Landmark Localization}

We formulate landmark localization as a ranked landmark prediction task rather than direct coordinate regression. Predicting multiple closest landmarks provides anatomical context, which improves robustness by leveraging the prior anatomical knowledge encoded in MLLMs. 

We pair each X-ray image with several annotated skeletal landmarks, ordered according to their distance from the image center. By introducing such image–landmark pairs spanning the entire body to the MLLM, the model learns to associate visual patterns with anatomical structures, thereby enabling spatial reasoning across diverse imaging views. 

Let $V_i$ denote the $i$-th CT volume, which has 14 annotated skeletal landmarks. The X-ray image $I_{i,j}$, corresponding to the $j$-th sample from $V_i$, is centered at position $P_{i,j}$. During supervised fine-tuning (SFT), the model $f_\theta$ is trained to generate an ordered set of skeletal landmarks associated with each image: $f_\theta(I_{i,j}) = \{ L_{i,k_1^\ast}, L_{i,k_2^\ast}, L_{i,k_3^\ast} \}
$, where $L_{i,k}$ denotes the $k$-th anatomical landmark of CT volume $V_i$, with $k \in \{1,\ldots,14\}$. The indices $\{k_1^\ast, k_2^\ast, k_3^\ast\}$ correspond to the three landmarks closest to $P_{i,j}$ in Euclidean distance. The predicted landmarks are ordered by proximity and expressed in a structured textual format consistent with the dataset format.

\subsection{Fine-tuning}
For efficient fine-tuning, we adopted the following strategies: \textbf{(1)} Unsloth Framework~\cite{unsloth}: is an open-source system designed to improve the efficiency and memory utilization of LLM fine-tuning. \textbf{(2)} Quantized Low-Rank Adaptation (QLoRA)~\cite{qlora}: keeps the original model weights frozen while introducing small trainable low-rank adapters within target layers, significantly reducing memory usage while retaining general knowledge~\cite{lora_vs_full}. 

The model $f_\theta$ was trained to generate a textual target conditioned on the corresponding visual input $I_{i,j}$, following the standard SFT cross-entropy loss below,
\begin{equation}
\mathcal{L}_{\text{SFT}} 
= - \mathbb{E}_{(I_{i,j}, T_{i,j}) \sim \mathcal{D}_{\text{sampled}}} 
\sum_{t=1}^{|T_{i,j}|} 
\log p_\theta \big( T_{i,j,t} \mid I_{i,j}, T_{i,j,<t} \big)
\label{eq:sft_loss}
\end{equation}
where $T_{i,j,t}$ denotes the $t$-th token in the output sequence, $\mathcal{D}_{\text{sampled}}$ denotes the sample data distribution, and $T_{i,j,<t}$ represents the preceding tokens. Full prompt templates, training scripts, and configuration files are provided in our public repository.

\subsection{C-arm Navigation}

We also conduct a qualitative experiment to demonstrate the reasoning capability and spatial awareness of the fine-tuned MLLM in C-arm navigation. Effective C-arm control requires accurate perception of the current spatial localization and reasoning about how the view should be adjusted to reach the target. To evaluate this capability, we design a navigation task based on chain-of-thought (CoT) prompting~\cite{chain_of_thought}.

Let $x_0$ and $r_0 = \varnothing$ denote the initial C-arm's field-of-view (FoV) X-ray image and response respectively. At step $t$, the model $f_\theta$ receives the previous FoV X-ray $x_{t-1}$ and the previous model response $r_{t-1}$ as input,  and generates a structured response $r_t$.
The response consists of two stages. In the first stage, the model identifies the closest anatomical landmark $\hat{\ell}_t$. In the second stage, conditioned on the predicted landmark, the model reasons about the required C-arm movement in anatomical space and outputs a discrete 2D motion command as follows:
\begin{equation}
a_t = (d_t^{x}, e_t^{x}, d_t^{y}, e_t^{y}), \quad
\left\{
\begin{aligned}
d_t^{x} &\in \{\text{LEFT}, \text{CENTER}, \text{RIGHT}\}, \\
d_t^{y} &\in \{\text{UP}, \text{CENTER}, \text{DOWN}\}, \\
e_t^{x}, e_t^{y} &\in \{\text{NONE}, \text{SMALL}, \text{MODERATE}, \text{LARGE}\},
\end{aligned}
\right.
\end{equation}
where the movements $e_t^{x}, e_t^{y}$ correspond to the step magnitude $\{0, 30, 60, 90\}$ millimeters, and $d_t^{x}, d_t^{y}$ refer to the direction. The landmark prediction and motion reasoning are generated together and returned as an XML-style response $r_t = f_{\theta}(x_{t-1}, r_{t-1})$. 
Based on the predicted motion command $a_t$, the C-arm pose is updated from $p_{t-1}$ to $p_t$, and a new radiograph $x_t$ is generated from $p_t$ using DeepDRR~\cite{deepdrr}. This process is repeated iteratively, allowing the model to reference its prior reasoning and actions. Overall, the perception-action loop is summarized below,
\begin{equation}
    r_t = f_\theta(x_{t-1}, r_{t-1}), \qquad
    x_t = \mathrm{DeepDRR}(p_{t-1}, a_t)
\end{equation}

\section{Experiments}
\label{sec:result}
\subsection{Synthetic and Real Data}
\label{sec:dataset_construction}
\begin{figure}[!t]
  \centering
  \caption{Illustration of the synthetic data construction. (a) As can be seen in the \textit{Sampling Summary}, C-arm isocenters is densely sampled within each CT volume: The superior–inferior axis is uniformly sampled over 70\% of anatomical height. The left–right axis is sampled from a Gaussian distribution centered on the anatomy with standard deviation equal to the average inter-humeral head distance (285~mm). The anterior–posterior axis is sampled from a broader Gaussian distribution ($\sigma = 100~\text{mm}$) to vary in magnification. For each sampled C-arm isocenter, a DRR is generated using DeepDRR~\cite{deepdrr}. (b) The \textit{landmark list} illustrates the 14 annotated skeletal landmarks. (c) The \textit{dataset summary} provides an example of a single X-ray image with three annotated landmarks, shown in the context of the full upper body.}
  \label{fig:Fig2}
    \begin{subfigure}[b]{0.25\textwidth}
        \includegraphics[width=\textwidth]{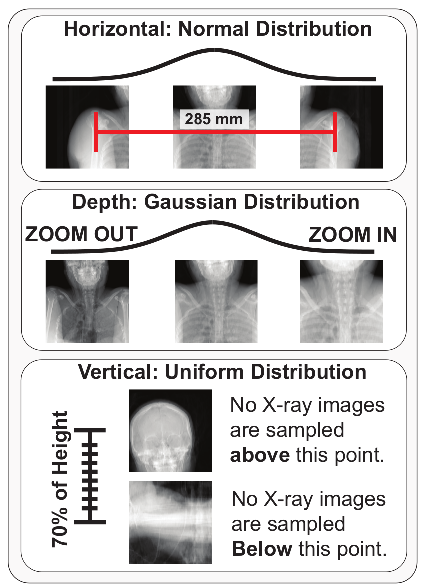} 
        \caption{Sampling Summary}
        \label{fig:Fig2_a}
    \end{subfigure}
    \hfill
    \begin{subfigure}[b]{0.46\textwidth}
        \includegraphics[width=\textwidth]{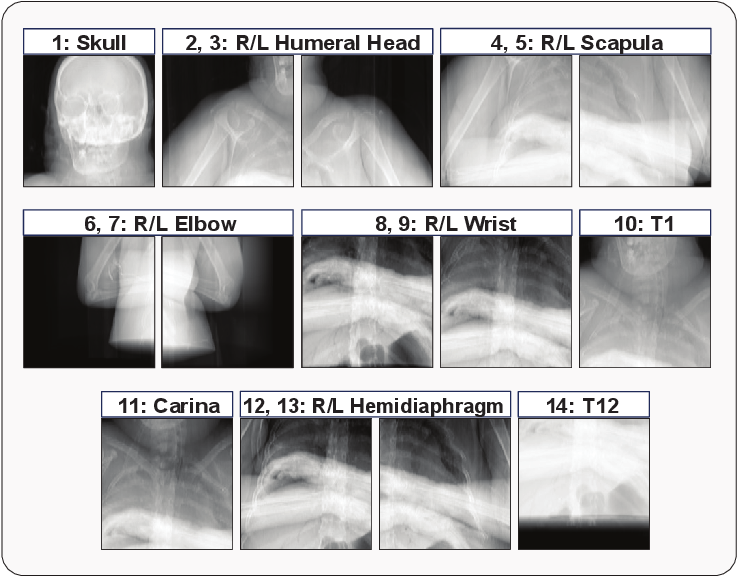}
        \caption{Landmark List}
        \label{fig:Fig2_b}
    \end{subfigure}
    \hfill
    \begin{subfigure}[b]{0.27\textwidth}
        \includegraphics[width=\textwidth]{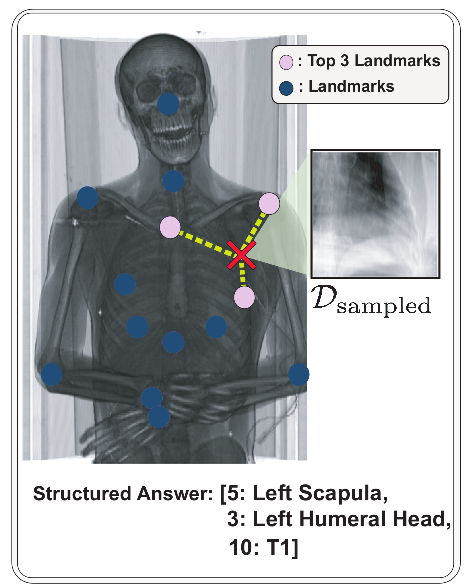}
        \caption{Dataset Summary}
        \label{fig:Fig2_c}
    \end{subfigure}
\end{figure}

We construct a synthetic X-ray dataset paired with skeletal landmark labels for fine-tuning MLLMs. These X-ray images are digitally reconstructed radiographs (DRRs) generated with DeepDRR~\cite{deepdrr} from CT volumes obtained from the New Mexico Decedent Image Dataset (NMDID)~\cite{NMDID}. We chose DeepDRR for its efficiency and realistic modeling of C-arm imaging geometry, such as isocenter. All coordinates in the generated images are expressed in millimeters (mm).

We utilize the annotated dataset introduced in~\cite{ISBI_carm}, where fourteen anatomical landmarks were defined across major upper-body skeletal structures and annotated using a custom graphical interface by domain experts. For each CT volume~$i$, the $k$-th landmark is denoted as $L_{i,k} = (x_{i,k}, y_{i,k}, z_{i,k})$, where $k \in \{1,\ldots,14\}$; we refer to this landmark dataset as $\mathcal{D}_{\text{annotated}}$. To generate diverse X-ray projections, we adopt the sampling strategy proposed in~\cite{ICCV_carm} to densely sample C-arm isocenters within each CT volume (see~\cref{fig:Fig2_a}). This sampling produces a synthetic dataset $\mathcal{D}_{\text{sampled}}$, where each sampled isocenter is represented as $P_{i,j} = (x_{i,j}, y_{i,j}, z_{i,j})$. For each sampled C-arm isocenter $P_{i,j} \in \mathcal{D}_{\text{sampled}}$, we compute the Euclidean distance $d_{i,j,k}$ to every annotated anatomical landmark $L_{i,k} \in \mathcal{D}_{\text{annotated}}$. To capture local anatomical context, each sampled position $P_{i,j}$ is associated with its three nearest landmarks. Formally, the set $\{L_{i,k^\ast}\}_{k=1}^{3}$ consists of the three landmarks in $\mathcal{D}_{\text{annotated}}$ with the shortest distances $d_{i,j,k}$ to $P_{i,j}$, ordered from closest to farthest. This ordered set defines a local anatomical reference frame that provides spatial cues encoding both positional and orientational information, and is formatted as \texttt{[index1: landmark1, index2: landmark2, index3: landmark3]}. To enhance training robustness and linguistic variability, each \texttt{landmark} name is randomly sampled during training from a set of three to four semantic variants (e.g., Skull, Cranium, Cranial vault, Calvarium). Overall, this sampling procedure yields 51,200 image–answer pairs for training and 10,240 for testing, obtained by sampling 1,024 projections per CT volume across 50 training and 10 testing volumes. Figure~\ref{fig:Fig2} provides illustrations of the synthetic dataset. 

Although our primary experiments are conducted on synthetic data, we additionally evaluate the proposed framework on real clinical data to assess potential sim-to-real gaps. The real X-ray dataset~\cite{unifesp} comprises 1,738 X-ray images annotated with 22 anatomical landmarks in a multi-label setting. These landmarks are designed to cover various body parts (e.g., the hand, ankle, cervical spine, etc.). A total of 1{,}564 images are used for fine-tuning, with the remaining 174 reserved for testing.

\subsection{Implementation Detail}
\textbf{Backbones:} We evaluate two families of multimodal large language models (MLLMs): Gemma3~\cite{gemma_2025} (4B, 12B, and 27B parameters), its medically fine-tuned variant MedGemma~\cite{MedGemma} (4B and 27B), and Qwen2.5-VL~\cite{qwen2.5-VL} (7B and 32B). Multiple parameter scales are evaluated within each family to study the effect of model size. In addition, Gemma3 is compared with MedGemma to assess the impact of domain-specific medical fine-tuning.

All fine-tuning using synthetic data is conducted on a single NVIDIA H200 GPU for three epochs. Quantized Low-Rank Adaptation (QLoRA) adapters are applied to the vision, language, and attention layers, with rank $r=16$ and scaling factor $\alpha=32$. When fine-tuning on real data, the same training configuration is adopted, with the exceptions that the models are trained for five epochs and additional LoRA adapters are introduced in the MLP modules to mitigate data scarcity.

\noindent\textbf{Baseline:} As a conventional deep learning baseline, we train the model proposed in~\cite{ISBI_carm} using the same synthetic training dataset $\mathcal{D}_{\text{sampled}}$. This baseline is trained to classify the single closest anatomical landmark given a sampled X-ray image $I_{i,j}$. During inference on the synthetic test set, a softmax function is applied over the predicted landmark classes and the top-3 predictions are extracted. In contrast, for real X-ray images, a sigmoid function is applied and the top-2 predictions are selected. The baseline model is trained for 25 epochs using the same data split.

\subsection{Skeletal Landmark Localization Results}

\begin{table}[!b]

\centering
\caption{Precision@K and Recall@K for closest landmark retrieval, grouped by model family, parameter size, and model variant (Var), with a deep learning (DL) baseline.
The optimal scores are 1.0 for Precision@K and 0.33, 0.66, and 1.0 for Recall@K at $K=1,2,3$, respectively. For each metric, the \textbf{bolded} value indicates the best-performing model, while the \underline{underlined} value indicates the second-best-performing model. MMLU-Pro~\cite{mmlu_pro} results are reported as the percentage change from the base model. It is evident that all fine-tuned models still possess their conversational abilities, as no drastic change was reported.}
\label{tab:Tab1}
\begin{tabular}{lllccc|ccc|c}

\toprule
\textbf{Model} & \textbf{Size} & \textbf{Var} &
\multicolumn{3}{c}{\textbf{Precision@K}} &
\multicolumn{3}{c}{\textbf{Recall@K}} &
MMLU-Pro \\
\cmidrule(lr){4-6}\cmidrule(lr){7-9}
 &  &  & @1 & @2 & @3 & @1 & @2 & @3 & $\Delta$\% \\
\midrule
\textbf{DL}~\cite{ISBI_carm}&21M& N/A&\textbf{0.995}&0.91&0.76&\textbf{0.332}&\underline{0.61}&0.76&N/A \\
\midrule
\multirow{8}{*}{\textbf{Gemma-3}}
 & \multirow{3}{*}{27B}
 & Base       & 0.64 & 0.61 & 0.54 & 0.21 & 0.41 & 0.54 & -- \\
 &  & Med & 0.57 & 0.52 & 0.49 & 0.19 & 0.35 & 0.50 & $\downarrow$\,4.91 \\
 &  & FT       & \underline{0.96} & \underline{0.92} & \underline{0.84}
                & \underline{0.32} & \underline{0.61} & \underline{0.84}
                & $\downarrow$\,1.04 \\
\cmidrule(lr){2-10}

 & \multirow{2}{*}{12B}
 & Base       & 0.58 & 0.54 & 0.47 & 0.19 & 0.36 & 0.47 & -- \\
 &  & FT       & \underline{0.96} & \textbf{0.93} & \textbf{0.85}
                & \underline{0.32} & \textbf{0.62} & \textbf{0.85}
                & $\uparrow$\,0.54 \\
\cmidrule(lr){2-10}

 & \multirow{3}{*}{4B}
 & Base       & 0.43 & 0.36 & 0.40 & 0.14 & 0.24 & 0.40 & -- \\
 &  & Med & 0.45 & 0.36 & 0.44 & 0.15 & 0.24 & 0.44 & $\downarrow$\,5.31 \\
 &  & FT       & \underline{0.96} & 0.91 & 0.82
                & \underline{0.32} & \underline{0.61} & 0.82
                & $\downarrow$\,1.87 \\

\midrule
\multirow{4}{*}{\textbf{Qwen-2.5VL}}
 & \multirow{2}{*}{32B}
 & Base       & 0.58 & 0.52 & 0.48 & 0.19 & 0.34 & 0.48 & -- \\
 &  & FT       & \underline{0.96} & \underline{0.92} & 0.83
                & \underline{0.32} & \textbf{0.62} & 0.83
                & $\downarrow$\,2.25 \\
\cmidrule(lr){2-10}

 & \multirow{2}{*}{7B}
 & Base       & 0.46 & 0.42 & 0.37 & 0.15 & 0.28 & 0.37 & -- \\
 &  & FT       & 0.95 & \underline{0.92} & 0.82
                & \underline{0.32} & \underline{0.61} & 0.82
                & $\uparrow$\,0.02 \\
\bottomrule
\end{tabular}
\end{table}

\noindent\textbf{Closest Landmark Retrieval:} To evaluate the effectiveness of fine-tuning, we conduct a top-3 closest landmark retrieval task on a test set of image-landmark pairs. Precision@K (i.e., accuracy of the highest-ranked predictions) and Recall@K (i.e., coverage of relevant items) are used as evaluation metrics. For a set of ground-truth relevant landmarks $G$ and the set of top-$K$ predictions $P_K$, we define: $\text{Recall@}K = \frac{|P_K \cap G|}{|G|},
\text{Precision@}K = \frac{|P_K \cap G|}{K}$. 

As shown in~\cref{tab:Tab1}, the fine-tuned models consistently outperform both the base model and MedGemma~\cite{MedGemma} across all values of $K$. This demonstrates that fine-tuning improves both landmark coverage R@$K$ and ranking accuracy P@$K$. Notably, the fine-tuned models achieve performance that is comparable to or better than the deep learning baseline for $K \ge 2$, whereas the baseline achieves near-perfect score in the $K = 1$ setting.

To assess whether domain-specific fine-tuning affected general model capabilities, we evaluate all models on the MMLU-Pro benchmark~\cite{mmlu_pro}. As shown in~\cref{tab:Tab1}, the fine-tuned models achieve performance comparable to their base counterparts, with score difference of less than 2\% compared to their base models, indicating that general language understanding and reasoning abilities are preserved.

\begin{figure}[!b]
  \centering
  \caption{(a) Landmark classification results shown as heatmaps. A prediction is considered correct if the ground truth landmark is within the top 2 predictions. As evidenced by the heatmaps, fine-tuning results in significant improvements. (b) In this landmark localization task, the ground-truth landmark for the X-ray is the left elbow (7). Both the base and fine-tuned Gemma 3 12B models initially produce incorrect predictions. After receiving clinician feedback prompting a re-evaluation, only the fine-tuned MLLM is able to produce a correct adjustment through reasoning, demonstrating that spatial awareness is acquired via fine-tuning.
  }
  \label{fig:Fig3}
    \begin{subfigure}[b]{0.3\textwidth}
      \includegraphics[width=\textwidth]{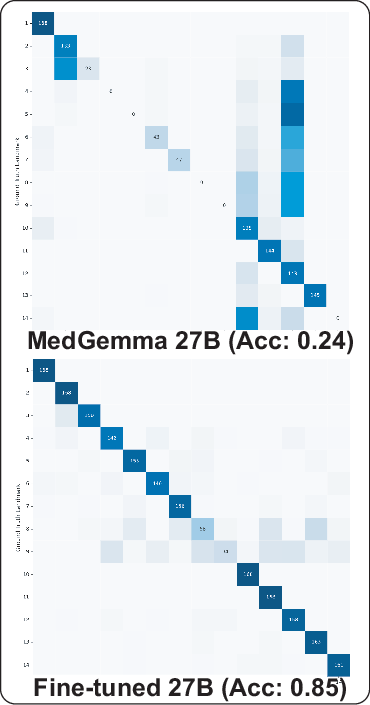} 
      \caption{Per Landmark Heatmap}
      \label{fig:Fig3_a}
  \end{subfigure}
  \hfill
  \begin{subfigure}[b]{0.69\textwidth}
      \includegraphics[width=\textwidth]{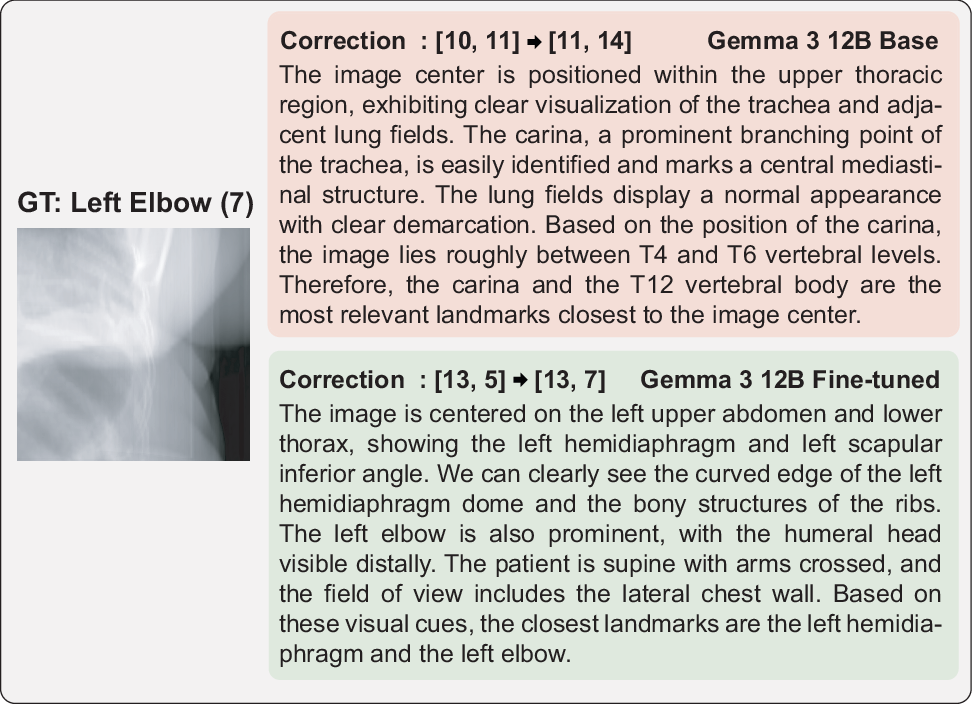} 
      \caption{Correction Reasoning}
      \label{fig:Fig3_b}
  \end{subfigure}
\end{figure}

\begin{table}[htbp]
\centering
\caption{Hit@2 and Hit@1 on synthetic and real X-ray datasets, grouped by model family, parameter size, and model variant (Var). For each metric, the \textbf{bolded} value indicates the best-performing model, while the \underline{underlined} value indicates the
second-best-performing model}
\label{tab:Tab2}
\begin{tabular}{lllcc|cc}
\toprule
\textbf{Model} & \textbf{Size} & \textbf{Var} &
\multicolumn{2}{c}{\textbf{Synthetic X-ray}} &
\multicolumn{2}{c}{\textbf{Real X-ray}} \\
\cmidrule(lr){4-5}\cmidrule(lr){6-7}
 &  &  & H@2 & H@1 & H@2 & R@2 \\
\midrule
\textbf{DL}~\cite{ISBI_carm} & 21M & N/A
& \textbf{0.92} & \textbf{0.81} & \textbf{0.95} & \textbf{0.94} \\
\midrule
\multirow{8}{*}{\textbf{Gemma-3}}
 & \multirow{3}{*}{27B}
 & Base        & 0.42 & 0.28 & 0.90 & 0.86 \\
 &  & Med   & 0.24 & 0.16 & 0.74 & 0.72 \\
 &  & FT         & \underline{0.85} & 0.73 & \underline{0.93} & \underline{0.91} \\
\cmidrule(lr){2-7}
 & \multirow{2}{*}{12B}
 & Base        & 0.41 & 0.26 & 0.86 & 0.82 \\
 &  & FT         & 0.84 & \underline{0.74} & 0.88 & 0.85 \\
\cmidrule(lr){2-7}
 & \multirow{3}{*}{4B}
 & Base        & 0.33 & 0.22 & 0.83 & 0.79 \\
 &  & Med   & 0.23 & 0.15 & 0.76 & 0.74 \\
 &  & FT         & 0.66 & 0.55 & 0.85 & 0.84 \\
\midrule
\multirow{4}{*}{\textbf{Qwen-2.5VL}}
 & \multirow{2}{*}{32B}
 & Base        & 0.33 & 0.21 & 0.79 & 0.75 \\
 &  & FT         & 0.70 & 0.61 & 0.88 & 0.86 \\
\cmidrule(lr){2-7}
 & \multirow{2}{*}{7B}
 & Base        & 0.32 & 0.20 & 0.69 & 0.65 \\
 &  & FT         & 0.65 & 0.59 & 0.83 & 0.81 \\
\bottomrule
\end{tabular}
\end{table}

\noindent\textbf{Skeletal Landmark Localization: }We further evaluated the models in a controlled setting where the target anatomical landmark is centered in the X-ray image. This experiment assesses whether the models can correctly associate a visual cue with its corresponding anatomical landmark. Given a centered X-ray image from $\mathcal{D}_{\text{annotated}}$, the model is prompted to predict the two most likely landmarks. Performance is evaluated using Hit@K (K = 1, 2) on the synthetic X-ray dataset, and Hit@2 and Recall@2 on the real X-ray dataset~\cite{unifesp}. Hit@K indicates whether at least one ground-truth landmark appears among the top-$K$ model predictions.

As shown in~\cref{tab:Tab2}, the fine-tuned models exhibit significant performance improvements over both their base and medically fine-tuned variants (MedGemma~\cite{MedGemma}) on both synthetic and real X-ray data. In particular, the fine-tuned Gemma-3 27B model achieves competitive performance (H@2: 0.85, H@1: 0.74) compared to the DL baseline proposed in~\cite{ISBI_carm} (H@2: 0.92, H@1: 0.81) in synthetic X-ray.

Per-landmark heatmap analysis, shown in~\cref{fig:Fig3_a}, provides further insight into model behavior. MedGemma 27B~\cite{MedGemma} frequently misclassifies a wide range of landmarks as T1 (10) or Right Hemidiaphragm (12), indicating a strong bias toward chest-related anatomical structures. This bias is substantially reduced in the fine-tuned models, suggesting that the proposed fine-tuning strategy improves landmark discrimination beyond chest X-ray. Additionally, we observed that misclassifications in the fine-tuned models are primarily concentrated around the elbow (6, 7) and wrist (8, 9), where there is substantial landmark overlap in arm-crossed acquisition settings.

In addition, we conduct a qualitative correction experiment to assess whether the MLLM exhibits reasoning capability and spatial awareness. In this setting, the model is prompted to revise an initially incorrect landmark prediction. The example in~\cref{fig:Fig3_b} shows that the fine-tuned model can leverage improved spatial anchoring to correct anatomically ambiguous predictions, highlighting its potential for agentic interaction with clinicians during surgery, a capability that conventional deep-learning approaches and the base Gemma 3 lack.

\subsection{Preliminary C-arm Navigation Results}

\begin{figure}[!t]
  \centering
  \caption{Qualitative C-arm navigation results using the fine-tuned Gemma-3 12B model. Given that our application targets neurovascular surgery, the C-arm is typically directed toward the patient’s head. Accordingly, in the left image, we select the left scapula as the starting position and the skull as the target position. In the right image, the C-arm navigation starts from the right scapula and targets the skull. At each step, the model identifies the closest anatomical landmark and predicts an action to move the C-arm toward the target. }
  \label{fig:Fig4}
  \includegraphics[width=\textwidth]{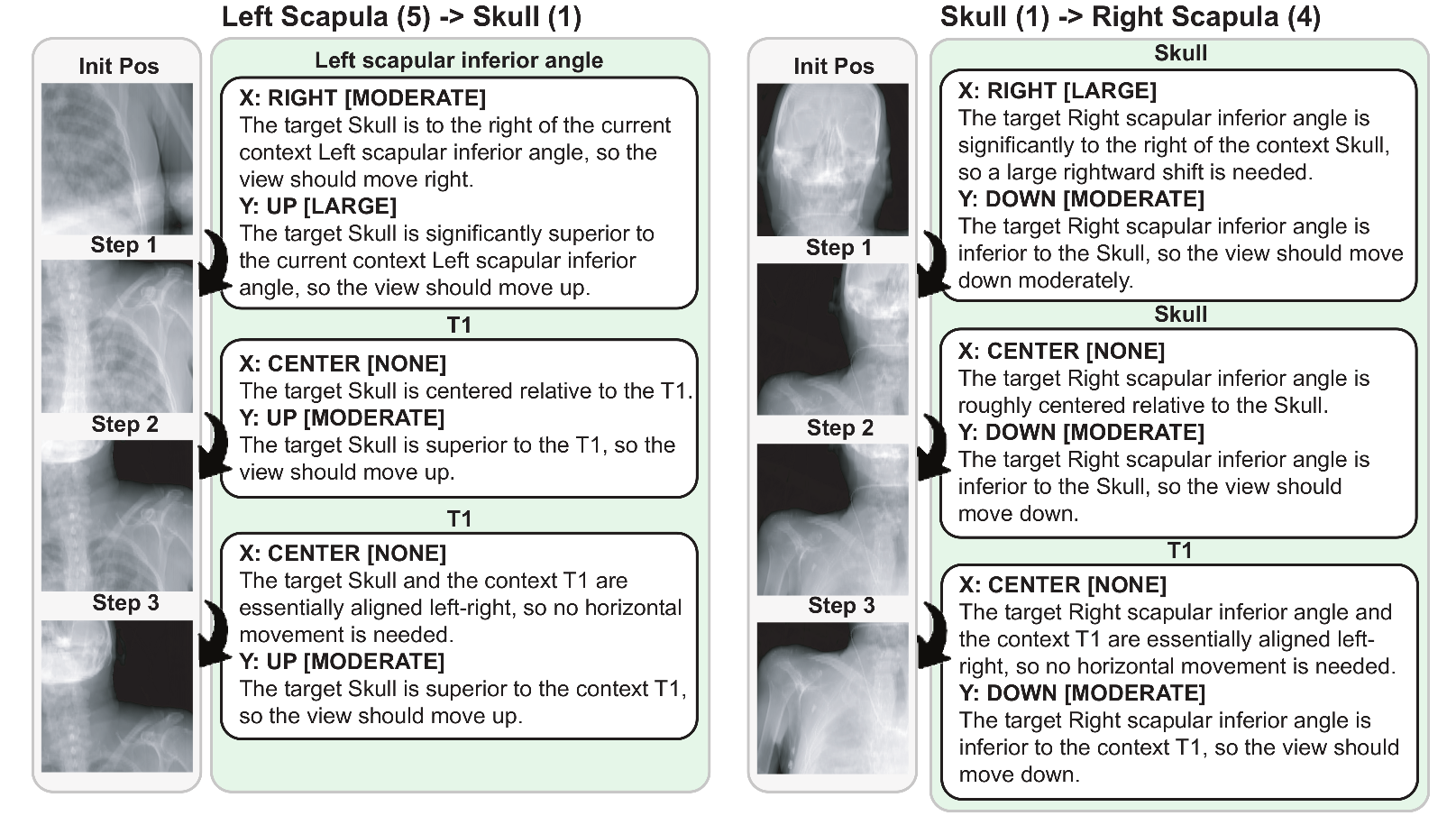}
\end{figure}

To investigate the potential for agentic C-arm control, we performed a qualitative C-arm perception–action loop experiment, initializing the C-arm at predefined anatomical landmarks and tasking it with reaching specified target landmarks.

Given that our application targets neurovascular surgery, the C-arm is typically directed toward the patient’s head. Here, we test whether the fine-tuned MLLM can navigate the C-arm from the right scapula to the skull, as well as in the reverse direction. Navigation was performed using a multi-step perception–action loop, as described in~\cref{sec:methods}. Figure~\ref{fig:Fig4} illustrates that the fine-tuned model iteratively identifies nearby anatomical landmarks and applies spatial reasoning to predicts an action to move the C-arm toward the target.

\section{Conclusion and Discussion}
\label{sec:conclusion}

We investigated the skeletal landmark localization performance of fine-tuned multimodal large language models (MLLMs) on X-ray images. After fine-tuning, the MLLMs demonstrated consistent improvements in localization accuracy (Tables~\ref{tab:Tab1} and~\ref{tab:Tab2}) compared to their base models. While conventional deep learning (DL) approaches~\cite{ISBI_carm} achieve slightly lower localization performance, the fine-tuned MLLMs attain competitive results on the evaluation benchmarks. More importantly, through qualitative experiments, we provide clear evidence of the reasoning ability and spatial awareness of the fine-tuned MLLMs. Specifically, the models can incorporate clinician feedback and perform adjustments when the initial localization is incorrect (\cref{fig:Fig3_b}), and can navigate the C-arm toward a target location in a multi-step manner (\cref{fig:Fig4}).

At the current stage, MLLMs may not yet match the localization precision of conventional DL models. A more practical approach is therefore to use a DL model for the initial prediction and leverage an MLLM to augment the DL output. Such a hybrid framework can incorporate clinician feedback and iteratively refine predictions through MLLMs' reasoning. In the long term, with continued advances in  MLLMs and fine-tuning strategies, we believe that agentic approaches have the potential to consistently surpass training-from-scratch DL models. Another potential direction for improving the current approach is the training strategy. At present, we use supervised fine-tuning to endow the MLLM with so-called “geometric reasoning,” where knowledge learned from landmark localization may partially transfer to C-arm navigation. In future work, we plan to explore reinforcement learning to directly train the MLLM in a perception–action manner.

\section{Statements and Declarations}
\noindent\textbf{Conflict of Interest} The authors declare that there is no conflict of interest regarding the publication of this paper. 

\noindent\textbf{Acknowledgment} This work was supported by the National Science Foundation under Grants No. 2218063.

\bibliography{sn-bibliography}

\end{document}